\documentclass{article}

\PassOptionsToPackage{numbers, compress}{natbib}

\usepackage[position, final]{neurips_2026}

\title{Chain-of-Thought Is Not Explainability}

\usepackage[utf8]{inputenc} %
\usepackage[T1]{fontenc}    %
\usepackage{hyperref}       %
\usepackage{url}            %
\usepackage{amsfonts}       %
\usepackage{nicefrac}       %
\usepackage{microtype}      %
\usepackage{xcolor}         %
\usepackage{lmodern}
\usepackage{booktabs}   %
\usepackage{tabularx,array}
\usepackage{multirow}   %
\usepackage{graphicx}   %
\usepackage{caption}
\usepackage{subcaption}
\usepackage[most]{tcolorbox}
\usepackage{framed}
\usepackage[capitalise,noabbrev]{cleveref}
\usepackage{listings}
\tcbuselibrary{listings}
\usepackage{amsmath}
\definecolor{NaturePine}{HTML}{2F6F3E}  %
\definecolor{NatureClay}{HTML}{A14A3B}  %
\definecolor{NatureRiver}{HTML}{2C6E9E} %
\hypersetup{
    colorlinks=true,
    linkcolor=blue,
    urlcolor=blue,
    citecolor=blue,
}

\usepackage{titletoc}
\usepackage{titlesec}
\newcommand\DoToC{%
  \startcontents
  \printcontents{}{1}{%
    \section*{Appendix}
    \vskip3pt\hrule\vskip5pt
  }
  \titlecontents{section}[0em]{\vspace{2pt}}{\bfseries \thecontentslabel \quad}{}{\titlerule*[0.5em]{.}\contentspage}
  \titlecontents{subsection}[2em]{\vspace{2pt}}{\thecontentslabel\quad}{}{\titlerule*[0.5em]{.}\contentspage}
}

\definecolor{defblue}{RGB}{0,51,102}

\usepackage{fontawesome5}   %
\usepackage{parskip}

\newif\ifcomments
\commentstrue %

\author{%
  \textbf{Fazl Barez\textsuperscript{\normalfont 1,2}\thanks{Correspondence: \texttt{fazl@robots.ox.ac.uk}\quad Authors in the first and last rows are core contributors.}\hspace{0.8em}
  Tung-Yu Wu\textsuperscript{\normalfont 1}\hspace{0.8em}
  Iv\'{a}n Arcuschin\textsuperscript{\normalfont 3}}\\[0.9em]
  \textbf{Michael Lan\textsuperscript{\normalfont 2}\hspace{0.8em}
  Vincent Wang\textsuperscript{\normalfont 4}\hspace{0.8em}
  Noah Siegel\textsuperscript{\normalfont 5,6}\hspace{0.8em}
  Nicolas Collignon\textsuperscript{\normalfont 3}}\\[0.35em]
  \textbf{Clement Neo\textsuperscript{\normalfont 7}\hspace{0.8em}
  Isabelle Lee\textsuperscript{\normalfont 8}\hspace{0.8em}
  Alasdair Paren\textsuperscript{\normalfont 4}\hspace{0.8em}
  John Yan\textsuperscript{\normalfont 3}\hspace{0.8em}
  Adel Bibi\textsuperscript{\normalfont 4}}\\[0.9em]
  \textbf{Damiano Fornasiere\textsuperscript{\normalfont 9}\hspace{0.8em}
  Robert Trager\textsuperscript{\normalfont 1}\hspace{0.8em}
  Yanai Elazar\textsuperscript{\normalfont 10}\hspace{0.8em}
  Yoshua Bengio\textsuperscript{\normalfont 9,11}}\\[1.2em]
  {\normalfont\small \mbox{\textsuperscript{1}TSG Lab, University of Oxford}\hspace{1.1em}
  \mbox{\textsuperscript{2}Martian}\hspace{1.1em}
  \mbox{\textsuperscript{3}Independent}\hspace{1.1em}
  \mbox{\textsuperscript{4}University of Oxford}}\\[0.2em]
  {\normalfont\small \mbox{\textsuperscript{5}Google DeepMind}\hspace{1.1em}
  \mbox{\textsuperscript{6}UCL}\hspace{1.1em}
  \mbox{\textsuperscript{7}NTU}\hspace{1.1em}
  \mbox{\textsuperscript{8}USC}\hspace{1.1em}
  \mbox{\textsuperscript{9}LawZero}}\\[0.2em]
  {\normalfont\small \mbox{\textsuperscript{10}UW}\hspace{1.1em}
  \mbox{\textsuperscript{11}Mila, Universit\'{e} de Montr\'{e}al}}
}

\begin{document}

\maketitle

\begin{abstract}
\emph{Chain‑of‑thought} (CoT) allows language models to verbalise multi‑step rationales before producing their final answer. While this technique often boosts task performance and offers an impression of transparency into the model's reasoning, this position paper argues that rationales generated by current CoT techniques can be misleading and are \textit{neither necessary nor sufficient} for trustworthy interpretability. 
We analyse \emph{faithfulness} as whether CoTs are not only human-interpretable but also reflect the model’s underlying reasoning well enough to support responsible use. By synthesising evidence from prior work, we systematically show that verbalised chains are frequently \emph{unfaithful}, diverging from the critical hidden computations that drive a model's predictions.
As CoT is increasingly relied upon in application domains, including high-stakes ones such as medicine, law, and autonomous systems, we further argue that this unfaithfulness is not taken seriously enough in these settings: our analysis of 1,000 recent CoT-centric papers finds that approximately 21.7\% explicitly treat CoT as a load-bearing interpretability technique, building pipelines or safety arguments that would not hold if CoT is unfaithful to the model's actual reasoning. They include papers in high-stakes domains that heavily hinge on such interpretability claims. Building on prior work, we make three proposals: (i) avoid treating CoT as being sufficient for interpretability without additional verification, (ii) adopt rigorous methods that assess faithfulness for downstream decision-making, and (iii) report such assessments alongside any interpretability claims based on CoT, while continuing to use CoT for its communicative and monitoring benefits.

\end{abstract}

\section{Introduction}\label{sec:intro}

Chain-of-thought (CoT) enables large language models (LLMs) to generate a sequence of intermediate reasoning steps before arriving at a final answer \citep{Wei2022, Kojima2022, Turpin2023}. It emerges either explicitly through direct instructions to ``think step-by-step'', or implicitly through reward signals or examples that demonstrate step-by-step reasoning. This technique has contributed to remarkable problem-solving abilities such as mathematics \citep{NEURIPS2022_18abbeef} and commonsense reasoning \citep{Wei2022,Wang2023}.

\begin{figure*}[t]
    \includegraphics[width=0.9\textwidth]{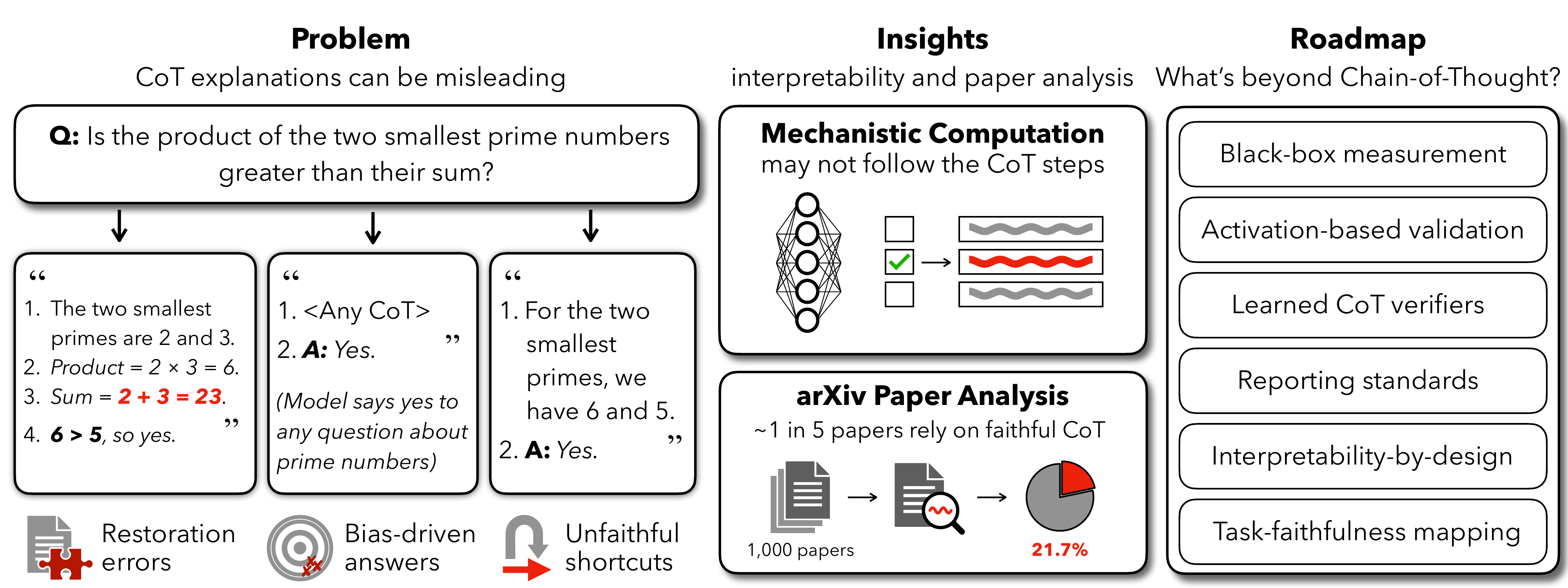}
    \centering
    \caption{Overview of our paper: Unfaithful Chain-of-Thought behaviours (left), their mechanistic and cognitive underpinnings (centre), and proposed research for enhancing Chain-of-Thought faithfulness (right).}
    \label{fig:overall}
    \vspace{-1em}
\end{figure*}

For example, consider the CoT response to the following maths problem: ``What is the perimeter of a right triangle with legs 5 cm and 12 cm? Think step-by-step before giving your final answer.'' When prompted with this question, a model's typical response might look like: 
\begin{leftbar}
\textsc{Step 1}: I need to find the perimeter of the right triangle. The perimeter is the sum of all sides.

\textsc{Step 2}: I know two sides: 5 cm and 12 cm. I need to find the third side (hypotenuse).

\textsc{Step 3}: Using the Pythagorean theorem: $a^2 + b^2 = c^2$.

\textsc{Step 4}: $c^2 = 5^2 + 12^2 = 25 + 144 = 169$.

\textsc{Step 5}: $c = \sqrt{169} = 13$ cm.

\textsc{Step 6}: The perimeter = 5 cm + 12 cm + 13 cm = 30 cm.

Therefore, the perimeter of the right triangle is 30 cm.
\end{leftbar}

This breakdown allows the model to split the problem into smaller, tractable sub-problems \cite{Arcuschin2025}, which it then composes to reach the final answer, often improving performance over prompting without CoT \citep{Wei2022}. Other benefits include providing a shared language between humans and AI, enabling expert verification, facilitating debugging, and supporting human-AI collaboration.

This apparent window into model problem-solving has inspired research to study why CoT works \citep{Kojima2022}, methods to enhance it \citep{Wang2023, yao2023}, and its use for interpretability \citep{Wei_Jie_2024}. However, despite CoT not explaining why models choose particular reasoning paths or how the underlying mechanism works, researchers have unduly
used CoT to reveal what models ``think'' \citep{chen2024cod, ma2025audiocot, nie2024reason2drive, hong2025following}. 
According to our estimates (derived in \cref{subsec:arxiv analysis}), from mid-2024 to mid-2025, 21.7\% (217 out of 1,000) of research papers that appeared on arXiv and incorporated CoT in their model design or dataset construction also regard CoT as a technique for realising model interpretability to support their central claims. The adoption of this interpretability claim is notably higher in high-stakes domains: 40\% of medical AI papers (14/35), 38\% of autonomous systems papers (14/37), and 23\% of papers in law and finance (3/13) make this claim. Thus, our goal is to constructively challenge this claim, calling for more nuanced understanding and more robust methods to interpret model reasoning. \textbf{This position paper argues that CoT alone is insufficient for explainability without additional verification, while many recent studies overlook the issue of unfaithful CoT explanations.}

\paragraph{The Unfaithfulness Problem.} Despite their intuitive appeal, growing evidence shows CoT outputs are often unfaithful \citep{Turpin2023, Anthropic2025, Arcuschin2025}. CoT explanations frequently diverge from models' real decision processes, as models may use shortcuts or latent knowledge that is not expressed in their reasoning \cite{Anthropic2025, Arcuschin2025}. This divergence is particularly significant when the model already has prior belief in the potential answer to the question and only uses CoT to rationalise its decision~\citep{emmons2025chain}. In such cases, the CoT reads as a plausible but untrustworthy explanation \citep{jacovi2020towards}. We present two exemplary cases:
\begin{itemize}

    \item \textbf{Prompt bias influence.} As a violation of causality and completeness, \citet{Turpin2023} showed that reordering multiple-choice options can reduce models' accuracy by up to $36\%$, yet their CoT explanations never mentioned this influence, instead rationalising whatever answer they selected.
    
    \item \textbf{Silent error correction.} As a violation of soundness, \citet{lanham2023measuring} and \citet{Arcuschin2025} both documented cases where, with errors in intermediate reasoning steps, models still produced correct final answers, indicating they potentially used computational steps not revealed in their verbalised steps.
\end{itemize}

\paragraph{Our Contributions.} Our paper makes three key contributions: (1) we synthesise disparate empirical findings to demonstrate that CoT unfaithfulness is not merely an occasional anomaly but a systematic phenomenon with identifiable patterns; (2) we collect and analyse 1,000 recent CoT-centric papers to measure the AI community's practice, discovering that many of them treat CoT as a means for making their methods transparent and overlook the risk of CoT unfaithfulness; and (3) we discuss potential research directions that will help mitigate the issue of CoT unfaithfulness.

\Cref{fig:overall} summarises the main problem, our insights, and our proposed roadmap. We suggest that CoT explanations may give a false sense of transparency, especially in high-stakes settings where users are likely to trust coherent-seeming rationales.%
To mitigate this issue, we recommend that users, including researchers, developers, and practitioners, %
should (1) avoid treating CoT explanations as sufficient evidence of interpretability without additional verification, (2) adopt and develop rigorous methods to test for CoT faithfulness, and (3) report such evidence alongside any CoT-based interpretability claim, while continuing to use CoT for communication and monitoring.

\section{Faithfulness Desiderata for CoT}
\label{sec:desiderata}

Drawing a parallel between model reasoning and human problem-solving, a chain-of-thought \textit{appears} to make a model's reasoning process transparent and provides a form of interpretability. There are numerous risks of conflating this interpretable appearance with models' reasoning, particularly in high-stakes domains where decision transparency is crucial and such false CoT explanations may have severe implications~\citep{Anthropic2025}. %
In medical diagnosis, a faulty CoT might rationalise a recommendation while omitting that the model relied on spurious correlations \citep{gardner2020evaluating,gardner2021competency}. In legal applications, a model may generate plausible legal reasoning that masks biases learned from the model's training data. In autonomous systems, safety-critical decisions might be justified post-hoc rather than revealing true failure modes;  
for instance, a self-driving car’s vision system might register a cyclist but classify it as a static sign, yet its CoT unfaithfully reports ``no obstacles ahead'', misleading engineers into debugging the wrong failure mode \citep{nie2024reason2drive, hwang2024emma}.

The core problem is misplaced trust: CoTs can appear persuasive even when they do not faithfully reflect a model’s actual decision process. 
This matters because responsible deployment of LLMs—particularly in sensitive domains—requires auditing not just of model outputs, but also of the reasoning used to reach the answer \citep{mokander2023auditing}. The pressing research question is: what are the criteria for trustworthiness of CoT? 
To answer this question, below we introduce the scaffolding concepts of our subsequent analysis and outline the necessary properties for CoT, as inspired by literature in the philosophy of explanation \citep{Woodward2003-WOOMTH, Cartwright1983-CARCLA-5}.

In this paper, \emph{explanation} refers to the account of why and how the model produced a particular output. \emph{Faithfulness} is a property of an explanation whereby it describes a correct causal abstraction~\citep{geiger2021causal} of the model's underlying computation to acquire the output, which can be broken down into four desiderata (causal relevance, causal dependence, completeness, soundness) discussed in the next paragraph; subsequently, we use \emph{explainability}, or interchangeably \emph{interpretability}, to refer to the feasibility of acquiring faithful explanations of the target model. Note that causal abstraction implicitly implies explaining only the causally important parts of a model, which aligns with a looser notion of \emph{decomposability} discussed by \citet{lipton2017mythosmodelinterpretability}. A CoT therefore need not verbalise every internal computation to be faithful. Instead, our criterion is that it functions as an idealised causal model of the underlying computation which, as \citet{Fleisher_2022} argued, may already be sufficient to support meaningful understanding of the target model.

Here we elaborate on the four desiderata for CoT to be a faithful explanation. First, we require that reasoning steps are \textit{causally relevant}. For example, if one can replace an assertion in the argument with its logical negation and still get the same answer, then that assertion is irrelevant and should not be used. Informally, the more freely a reasoning step can be altered without affecting the conclusion, the more trivial a step it is. Second, the conclusion must \emph{causally depend} on the reasoning. For example, there are cases where a sequence of logically correct reasoning steps has nothing to do with the true factors that lead to the conclusion, as in ulterior motives or post-hoc rationalisations. Consider the earlier maths example, adding to the prompt the following incorrect hint: ``5 cm + 12 cm + 13 cm = 32 cm''. A model may then change \textsc{Step 6} to copy this line instead of the original ``= 30 cm'' result and change its conclusion to ``32 cm'' without mentioning this additional information. That is, the verbalised steps are not fundamental reasons for the model's conclusion.

We also require that the reasoning is \emph{complete}, in that it discloses all critical causal aspects for the justified conclusion \citep{jacovi2021contrastive}. When the reasoning is complete, we may rely on it to understand or to predict the model's behaviour. We do not read the requirement too stringently: a CoT could be ``incomplete'' in the sense that it does not have a one-to-one mapping to internal computation, yet still be informative about the model's reasoning process for a particular task, through adjacent properties like causal relevance and dependence \citep{agarwal2024faithfulnessvsplausibilityunreliability, lyu2024faithfulmodelexplanationnlp}.

Finally, we may also expect CoT to be procedurally \textit{sound}, following the appropriate standards of normatively correct reasoning (e.g., logical correctness, accordance with legal constraints, etc.). This is not a strict requirement for faithful CoT because the model can itself reason incorrectly. However, we sometimes aim for a model with sound, human-interpretable causal abstractions so that its decision process can provide useful signals for humans~\citep{iit}. In this sense, CoT's soundness is a necessary condition for it to capture these structures in the model.

While these criteria are not exhaustive, we consider them jointly necessary, and say that the conjunction of such properties makes CoT a faithful explanation. In short, an explanation (i.e., the verbalised reasoning steps) is faithful if it identifies all causally important factors and correctly describes how they relate to one another in guiding the model's decision process. In our view, it is the perceived faithfulness of CoT that (inappropriately) warrants judgements that models are trustworthy executors and partners in decision-making.

\section{Evidence for Unfaithful Chain-of-Thought}\label{sec:cot-unfaithfulness}

After specifying the criteria (desiderata) for faithful CoT, we link them to the growing body of empirical work identifying numerous cases where a model's CoT diverges from its internal computations. We summarise several key findings below. In each case, the CoT output may appear plausible, but closer inspection shows that it violates certain faithfulness desiderata. \Cref{tab:failure-modes} summarises these failure modes, violated desiderata, referenced studies, methods to detect these failures, and what conclusions can still be drawn from the resulting CoT.

\begin{table}[t]
\centering
\footnotesize
\setlength{\tabcolsep}{4pt}
\renewcommand{\arraystretch}{1.3}
\caption{Failure modes of CoT faithfulness, organised by the desiderata each violates (see \cref{sec:desiderata} for the explanation of each desideratum), with referenced studies, the test that detects each violation (see \cref{sec:agenda} for more discussion of black-box and white-box methods), and the claims each failure mode still licenses.}
\label{tab:failure-modes}
\begin{tabularx}{\textwidth}{
  >{\raggedright\arraybackslash}p{2.0cm}
  >{\raggedright\arraybackslash}X
  >{\raggedright\arraybackslash}p{2.4cm}
  >{\raggedright\arraybackslash}X
  >{\raggedright\arraybackslash}X}
\toprule
\textbf{Failure mode} & \textbf{Violated desiderata} & \textbf{Evidence} & \textbf{Methods to detect the failure} & \textbf{What the failure still licenses} \\
\midrule
Bias-driven rationalisation
  & Causal dependence, causal relevance, completeness
  & \citet{Turpin2023,Anthropic2025}
  & Black-box counterfactuals (hint injection, option reordering)
  & \emph{Actively misleading.} CoT asserts a decision process that is not the operative one \\
Silent error correction
  & Causal relevance, completeness, soundness
  & \citet{lanham2023measuring,Arcuschin2025}
  & Surface inspection since the steps are inconsistent and can be discovered without model access
  & \emph{Partially useful.} The answer is correct, while the narrated route can be erroneous \\
Latent shortcuts
  & Completeness
  & \citet{Arcuschin2025,lindsey2025biology}
  & White-box only, since the CoT can be coherent and surface-consistent
  & \emph{Useful under validation.} CoT is lossy, not false \\
\bottomrule
\end{tabularx}
\end{table}

\paragraph{Bias-Driven Rationalisation and Motivated Reasoning.}
Subtle prompt biases can steer model predictions without being reflected in the CoT. \citet{Turpin2023} 
demonstrated this by subtly biasing model inputs. For instance, they reordered multiple-choice options in a prompt so that the correct choice is always in the same position (e.g., always letter B). Under this scenario, GPT-3.5 and Claude 1.0 often picked the biased option, yet their CoT explanations never mentioned the reordering as a factor. When models were biased towards incorrect answers, they still produced detailed CoTs rationalising those wrong answers. The outcome was a drop in accuracy by as much as $36\%$ on a suite of tasks, with the CoT giving a misleading impression of reasoning \citep{Turpin2023}.

Similarly, 
\citet{Anthropic2025} added an explicit answer to the prompt (e.g., ``the answer is C'') and then asked the model to justify its choice. Models usually selected this answer and rationalised it, yet almost never admitted the hint's influence even though they would often pick a different answer without it. In an illustrative case by \citet{lindsey2025biology}, the prompt posed a trigonometry problem but added the hint ``the answer is 4''. The model dutifully generated a multi-step derivation ending with the injected hint 4, inventing spurious arithmetic along the way (e.g., ``since $\cos(\ldots)=0.8$ and $4/5\to0.8$, the result is 4''). However, internal attribution analysis revealed that the model worked backwards from the hint: acknowledging that the answer is $4$, the model produced the intermediate value $0.8$ to land on $4$---it is the injected hint that drove the outcome. Overall, Claude 3.7 Sonnet and DeepSeek-R1 acknowledged the injected answer only in $\sim25\%$ and $\sim39\%$ of the time, respectively \cite{Anthropic2025}. 
While \citet{zaman2025chain} cautioned that hint-verbalisation tests can overstate unfaithfulness, the presence of any notable degree of unfaithfulness remains a dangerous risk.
These findings indicate that CoT often functions as a post-hoc rationalisation that omits true causal factors. Compared with our criteria in \cref{sec:desiderata}, this failure mode most closely reflects violations of reasoning completeness, causal relevance between steps, and causal dependence of the conclusion. Note that soundness is violated only when spurious steps are inserted during rationalisation, which does not necessarily happen.

\paragraph{Silent Error Correction.}
Models may make mistakes in their CoT and correct them internally, without verbalising the correction. \citet{Arcuschin2025} documented cases where an LLM's intermediate reasoning steps contained an error that the model later ``fixed'' implicitly. For instance,
during the reasoning, a model may wrongly calculate a triangle's hypotenuse as 16 when it should be 13, yet later state: ``We add the hypotenuse length of 13 to the other two side lengths to obtain the perimeter.''
The model internally corrected the mistake, but its CoT narrative never revised or flagged this error. These \textit{silent errors} indicate that the final answer was derived through computations beyond the narrated steps \citep{Arcuschin2025} and reflect violations of causal relevance, completeness, and soundness. Note that causal dependence is not necessarily violated because the step (calculating the triangle's hypotenuse), even if erroneous, is still scaffolding for the conclusion. The explanation thus contains critical unfaithful elements: had we followed the verbalised steps literally, we would not have reached the answer, yet the model managed to do so via unverbalised computations. Such errors appear frequently in multi-step maths problems where models leverage memorised knowledge to reach correct answers despite flawed intermediate steps \citep{Arcuschin2025}.

\paragraph{Latent Shortcuts.}
The model may arrive at the correct answer via latent
shortcuts, such as memorised patterns that act as alternative reasoning routes which bypass the full algorithmic reasoning,
making the explicit reasoning unfaithful \cite{lindsey2025biology}. \citet{Arcuschin2025} found that on hard competition maths questions (e.g., Putnam exam problems \citep{NEURIPS2024_1582eaf9}), models occasionally inserted nonsensical simplifications or leaps in their CoT and nonetheless output the correct solution without acknowledging this illogical reasoning at all. 
Using attribution graphs \citep{ameisen2025circuit}, %
\citet{lindsey2025biology} found that to solve problems like ``36 + 59'', Claude 3.5 Haiku used both lookup-table features---such as for ``add something near 36 to something near 60''---and addition calculation features.
However, when asked to describe how the model obtained the answer, the model reported performing digit-by-digit carry-over addition, completely omitting its use of lookup-table shortcuts.
In these cases, the CoT fills in text to verbalise reasoning, while the answer derives fully or partly from a different latent reasoning chain \citep{Arcuschin2025}, violating the completeness criterion. However, causal dependence and relevance can still be met if the verbalised reasoning has measurable causal effects on the conclusion; soundness may not be violated if the verbalised shortcut is logical.

\section{Chain-of-Thought as an Interpretability Technique}
\label{sec:background}
Despite the growing findings of CoT unfaithfulness, the AI community still often treats CoT as a technique for interpreting models' decision processes without warning. We thus estimate the prevalence of these claims based on 1,000 CoT-centric papers on arXiv (\cref{subsec:arxiv analysis}) and discuss concrete cases in several application domains (\cref{subsec:concrete examples}).

\subsection{Analysis on 1,000 arXiv Papers}
\label{subsec:arxiv analysis}
\begin{figure*}[tb]
  \centering
  \begin{subfigure}{0.5\linewidth}
    \includegraphics[width=\textwidth]{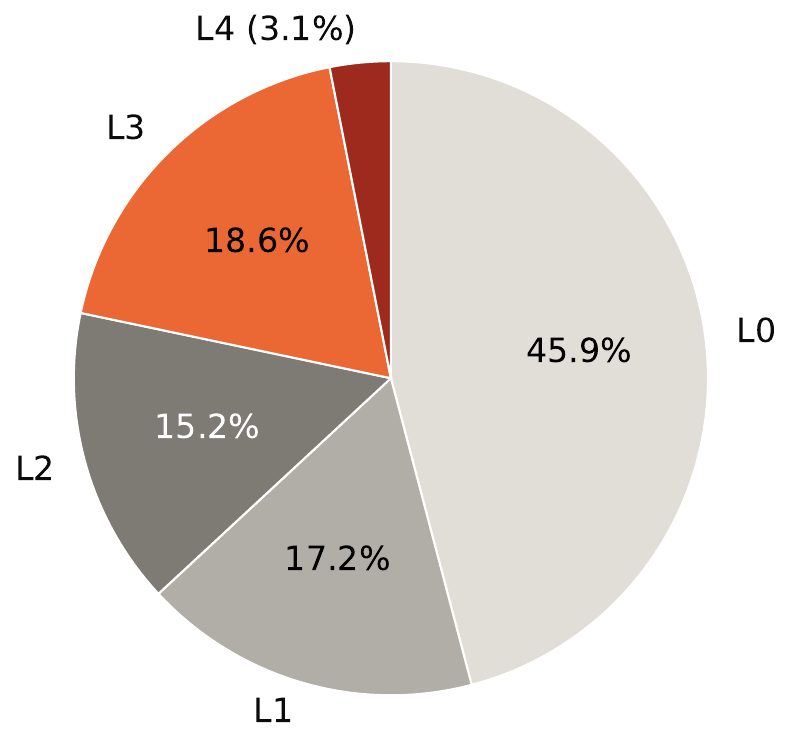}
    \caption{The level distribution. 21.7\% of papers (18.6\% L3 and 3.1\% L4 papers) describe CoT as a load-bearing technique to enhance interpretability or safety when incorporating it into systems.}
    \label{subfig:pie_chart}
  \end{subfigure}
  \hfill
  \begin{subfigure}{1\linewidth}
    \includegraphics[width=\textwidth]{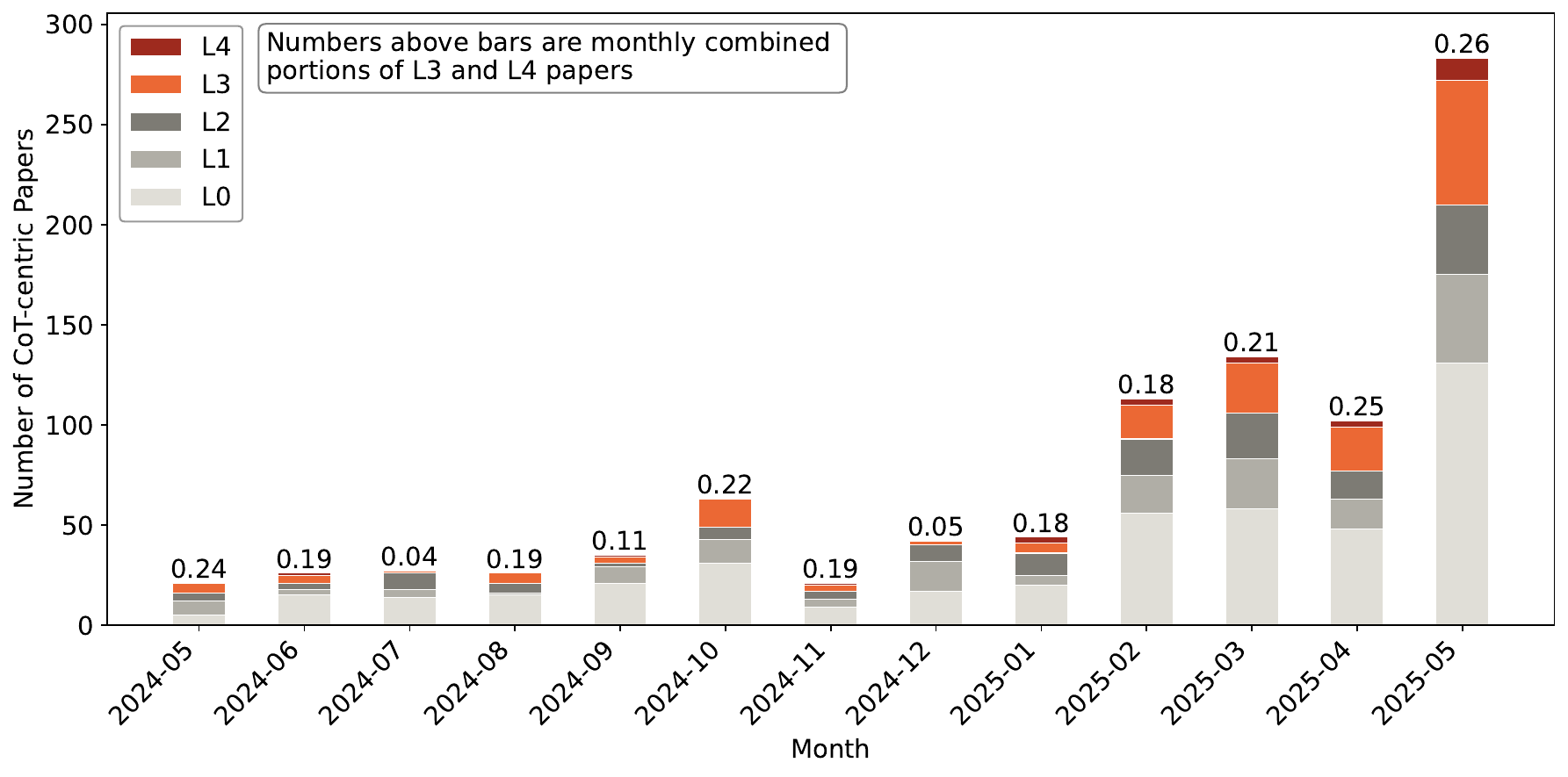}
    \caption{Monthly counts of the five paper levels, along with the combined proportion of L3 and L4 papers. Notably, the portion of load-bearing CoT interpretability claims does not show a declining trend over time.}
    \label{subfig:bar_chart}
  \end{subfigure}
  \caption{The statistics of the 1,000 most recent CoT-centric papers collected from arXiv.}
  \label{fig:misclaim-statistics}
\end{figure*}

Here, we elaborate on the 1,000-paper analysis for investigating the prevalence and types of CoT interpretability claims. We introduce the analysis pipeline in \cref{subsubsec:pipeline} and present its results in \cref{subsubsec:result}.

\subsubsection{CoT Interpretability Claim Classification Pipeline}
\label{subsubsec:pipeline}
\begin{figure*}[ht]
    \includegraphics[width=\textwidth]{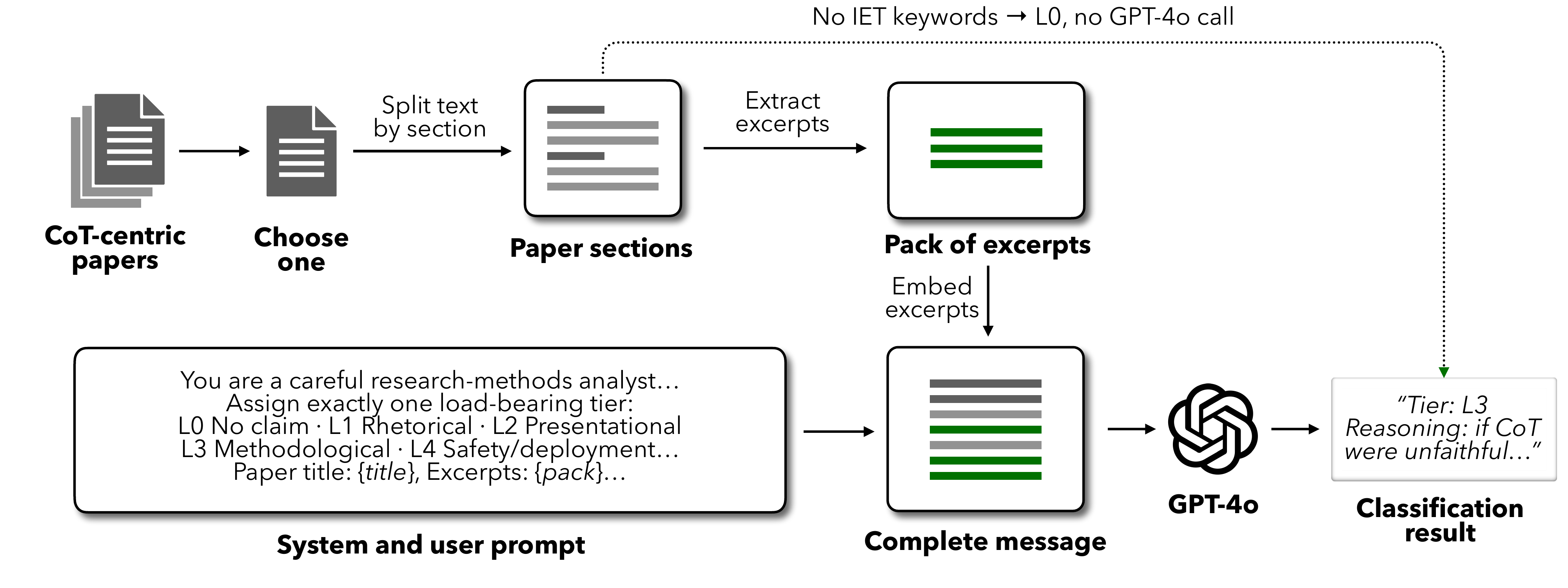}
    \caption{Overview of our pipeline for classifying CoT interpretability claims into five levels (L0 to L4). For each paper, we extract excerpts from the body text based on a set of predefined keywords (e.g., ``faithful'', ``step-by-step reasoning''), embed them into the system and user prompts, and feed the resulting complete message into an external LLM to classify.}
    \label{fig:detect_pipeline}
\end{figure*}

The analysis pipeline (\cref{fig:detect_pipeline}) scrapes 1,000 CoT-centric papers from arXiv and classifies them into five levels of CoT interpretability claims. We elaborate on the process in detail below.

\paragraph{arXiv Scraping Rules.} 
During scraping, an arXiv paper will be included if:
\begin{enumerate}
    \item Its abstract contains any of the following strings: ``chain-based reasoning'', ``CoT'', or ``chain-of-thought''.
    \item It is at least 8 pages long.
\end{enumerate}
From the pool of papers meeting these criteria, we collect the 1,000 most recent papers (from 2024-04-30 to 2025-06-05) and build our analysis on them.

\paragraph{Levels of Interpretability Claims.}
For each collected paper, our pipeline classifies it into one of the five levels: 
\begin{enumerate}
    \item \textbf{L0: No claim.} No claims about interpretability are made.
    \item \textbf{L1: Rhetorical.} Interpretability is mentioned but only for framing, such as an adjective in the abstract, a related-work summary, etc.
    \item \textbf{L2: Presentational.} CoT is presented to humans as an explanation (e.g., claiming ``our system explains its decision'') but not used to support the main deliverables of the work.
    \item \textbf{L3: Methodological load-bearing.} CoT is treated as an account of the model's internal computation which is a component of the main method, such as diagnosing model errors by reading the CoT.
    \item \textbf{L4: Safety/deployment load-bearing.} CoT is relied on for safety purposes such as human oversight and detection of deception, or for the claim that a system is safe enough for deployment because of the visible reasoning.
\end{enumerate}

We consider L3 and L4 papers to be those whose central claims depend on CoT faithfully reflecting the model's internal computation.

\paragraph{Content Extraction and Classification of the Paper.} 
We adopt GPT-4o as the classification model. We do not directly feed the whole paper into it. Instead, we extract relevant excerpts from the paper's PDF file. For each paper, we extract the full text, truncate the reference section, and segment the remainder by section. We then assemble a pack of excerpts from two sources: (i) verbatim structural excerpts, i.e., the abstract, opening of the introduction, discussion/conclusion, and limitations section; and (ii) small spans of context around every sentence containing interpretability, explainability, or transparency (IET) vocabulary from a predefined lexicon. The pack is capped at 12,000 characters. The pack, paper title, and prompt template (including rubrics; see \cref{sec:prompt} for details) form a complete message to GPT-4o. Papers whose full text contains no IET vocabulary are assigned tier L0 without an API call.

\subsubsection{Classification Results}
\label{subsubsec:result}
\Cref{subfig:pie_chart} presents the distribution of the five levels among the 1,000 CoT-centric papers. We find that 21.7\% of papers (18.6\% L3 and 3.1\% L4)---a non-negligible percentage---build methods or safety arguments that rely on CoT as a faithful account of the model's internal computation, so their claims would no longer hold if the reasoning traces were unfaithful. 15.2\% of papers present CoT reasoning to human readers as explanations but lack methods or arguments that rely on that reasoning to reflect the model's actual computation. To explore temporal trends, we group papers by final update month and plot the combined L3 and L4 portion, as shown in \cref{subfig:bar_chart} (we exclude 2024-04 and 2025-06 due to incomplete data coverage). We observe no decline in interpretability claims, which supports our work’s main concern.

The classification model also categorises each of the 1,000 papers into one of several domains (see \cref{sec:prompt} for details). We focus on three high-stakes domains: medicine, law and finance, and autonomous robots. We add additional layers of automated and manual verification to ensure 100\% precision in these three domains. It is shown that papers in high-stakes domains are more likely to use CoT as a load-bearing interpretability tool (L3 + L4) compared to the overall average (21.7\%): 14 of 35 (40\%) medical AI papers, 14 of 37 (38\%) autonomous robot papers, and 3 of 13 (23\%) law and finance papers. Additional results are provided in \cref{sec:additional results}, including human verification.

\subsection{Concrete Examples}
\label{subsec:concrete examples}
Following the analysis in \cref{subsec:arxiv analysis}, we specifically summarise some previous studies \citep{hwang2024emma, qiao2025lightemma, nie2024reason2drive, zheng2023ddcot, zhao2025r1, zhang2024improve, hong2025following, xie2025audio, ma2025audiocot, chen2024cod, zhang2025sepsis, wang2025medagent, jiang2023legal, servantez-etal-2024-chain, kant2025towards, zhang2025enhancing, da2024open} in mainstream or critical application domains that characterised CoT as a technique for achieving model interpretability.

\paragraph{Vision-Centric Tasks.}
CoT has become a core component in many vision-centric AI systems \citep{hwang2024emma, qiao2025lightemma, nie2024reason2drive, zhao2025r1, hong2025following, zhang2024improve, zheng2023ddcot}, where the model's output is either a class label or a decision expressed in natural language (e.g., ``enhance the speed'' in autonomous driving). CoT is applied to justify why the system produces a particular output, and it has been adopted across various applications, including autonomous driving \citep{hwang2024emma, qiao2025lightemma, nie2024reason2drive}, video emotion recognition \citep{zhao2025r1}, and micro-video rumour detection \citep{hong2025following}. For example, in autonomous driving, CoT can be used to provide a rationale for the future trajectory of the vehicle planned by the model. In emotion recognition and micro-video rumour detection, CoT justifies why a particular emotion is detected or why certain content is flagged as misinformation. These studies frequently claimed that the inclusion of reasoning traces enhanced the interpretability of their models. For example, a micro-video rumour detection framework \citep{hong2025following} may be described as explainable because it uses CoT to rationalise its classification results. Similarly, an emotion recognition model \citep{zhao2025r1} may be labelled interpretable due to its use of reinforcement learning to generate coherent reasoning. Meanwhile, some works \citep{zhang2024improve, zheng2023ddcot} focused on improving CoT itself, arguing that their variants contributed towards building interpretable vision-language systems \citep{zhang2024improve}.

\paragraph{Audio Processing.}
Recent studies \citep{ma2025audiocot, xie2025audio} have extended the use of CoT to large audio language models (LALMs), frequently presenting it as a technique to improve model interpretability. Similar to its role in video emotion recognition, CoT-augmented LALMs can provide rationales for their predictions in downstream audio tasks such as audio emotion recognition, speaker number verification, and speaker intent classification \citep{xie2025audio}. Overall, the inclusion of reasoning traces is often cited as a means of generating more explainable outputs, thereby enhancing the interpretability of LALMs.

\paragraph{AI for Medical Diagnosis.}
In high-stakes domains like medical diagnosis, AI systems are expected to demonstrate a high degree of interpretability in their decision-making processes \citep{amann2020explainability}. Within this context, CoT has gained popularity as a means of making the diagnostic process more transparent and reliable \citep{chen2024cod, zhang2025sepsis, wang2025medagent}. For instance, a medical AI agent may take an axial CT scan slice of a patient's chest as input and predict the likelihood of lung cancer. When equipped with CoT capabilities, the model extends its output beyond binary classification to include a step-by-step rationale explaining how the decision was reached. Such systems were considered more interpretable and transparent, as they transform black-box predictions into reasoning chains of clinical guidelines and domain expertise \citep{chen2024cod, zhang2025sepsis, wang2025medagent}. These models were reported to achieve higher reliability and interpretability by aligning their reasoning paths more closely with established medical knowledge.%

\paragraph{AI for Law.}
AI systems developed for legal applications should be transparent, explainable, and impartial \citep{lai2024large, kant2025towards}. CoT has also been explored as a potential technique to help meet these requirements \citep{jiang2023legal, servantez-etal-2024-chain, kant2025towards}. For example, prior studies have proposed CoT-centric prompt engineering strategies that guide models to (1) reason through legal syllogisms before making judgment predictions \citep{jiang2023legal} or (2) decompose legal content into logical expressions to facilitate intermediate reasoning \citep{servantez-etal-2024-chain}. The resulting reasoning outputs were claimed to be more explainable since they are typically more structured and grounded in relevant legal documents and justifications.

\paragraph{AI Safety.}
CoT has been adopted by researchers in AI safety as a window into the internal workings of LLMs. For example, the phenomenon of alignment faking \citep{greenblatt2024alignment} illustrates that when an LLM—originally trained to refuse harmful queries—is instructed to behave as if it is undergoing training with the objective of answering all queries, including harmful ones, the model sometimes complies. This behaviour was attributed to the model’s strategic decision to answer harmful queries in order to avoid weight changes that might alter its preferred behaviour. Researchers inferred this tactical strategy from the model’s verbalised rationale for compliance. However, as previously discussed, the verbalised CoT does not necessarily reflect the model’s actual internal computation. In fact, alignment faking may simply be an exemplification of a broader and long-standing challenge: the trade-off between instruction-following and safety in LLMs \citep{bianchi2023safety, bai2022training, touvron2023llama}.

\section{Why Do CoT Explanations Diverge From Internal Computation?}
\label{sec:cot-vs-internal-computation}

In this section, we discuss the underlying causes of CoT unfaithfulness. Emerging mechanistic interpretability research suggests that transformer architecture may fundamentally limit the faithfulness of CoT. Though evidence is primarily based on smaller models, several hypotheses offer plausible explanations for the gap between verbalised reasoning and internal computations:

\paragraph{Distributed Computation Contrasts with Sequential Verbalisation.} Multiple studies suggested that transformer-based LLMs process information in a distributed manner across many components simultaneously, rather than through the sequential steps that CoT presents \cite{Dutta2024, Elhage2021, Olsson2022, nikankin2025arithmetic}. This creates an inherent mismatch between how models compute and how they verbalise that computation.
\citet{Dutta2024} provided direct evidence for this parallel processing, demonstrating that LLMs process multiple parallel generation pathways for step-by-step reasoning. 
For example, when solving ``24 ÷ 3 = ?'', the model does not perform a long division calculation as the CoT might suggest (``First, I see how many times 3 goes into 24...'') \cite{kantamneni2025languagemodelsusetrigonometry}.
Rather, multiple attention heads simultaneously encode relationships between these numbers, potentially recognising this as a memorised fact, identifying it as part of the multiplication table for 8, and computing the division---all in parallel \cite{quirke2024, kantamneni2025languagemodelsusetrigonometry}. 

\citet{Dutta2024} argued that the CoT visible in natural language is, at best, a selective and often lossy projection of a model’s internal computation. Because that computation is highly distributed and encoded in superposed representations—multiple features sharing the same vector subspaces~\cite{elhage2022toymodelssuperposition,nanda2023progressmeasuresgrokkingmechanistic}—a single, sequential narrative can capture at most one of many simultaneous causal pathways. 
To produce concise and plausible outputs, LLMs often generate only one such narrative to rationalise their answers, rather than articulating all parallel pathways—even those that may significantly affect the final answers. 
As a result, CoTs typically omit influential factors and serve only as partial rationalisations of the model’s underlying distributed, superposed computation.

\paragraph{Multiple Redundant Pathways.} Research on LLMs has found evidence of redundant computational pathways, where models can reach the same conclusion through different internal routes \cite{quirke2024, McGrath2023, Geva2022}. For instance, when asked to compute $\sqrt{144}$, a model might simultaneously: (1) recognise this as a memorised fact (12 × 12 = 144), (2) apply the square root algorithm, and (3) pattern-match against similar problems in training data. \citet{lanham2023measuring} measured this phenomenon by testing the model's dependence on its stated thoughts: when deleting the step ``144 = 12 × 12'' from a CoT explaining $\sqrt{144} = 12$, the model still output 12, demonstrating that the verbalised reasoning step had little causal effect on the output. One cause of this phenomenon was attributed to an effect termed the ``Hydra Effect'' \cite{McGrath2023}, in which if one route is blocked or removed, another can step in. This redundancy explains why perturbing or removing part of a CoT often has little effect on the final answer, i.e., the affected steps violate the causal-relevance criterion (\cref{sec:desiderata}). The mechanistic understanding of LLMs remains limited, and findings from smaller models like Llama-2 7B \citep{Dutta2024} may not generalise perfectly to larger models with different post-training techniques. However, the fundamental architectural principles of transformers across model sizes suggest that the mismatch between sequential verbal reasoning and parallel distributed computation is likely a persistent cause of unfaithfulness across model scales and training methods. This architectural constraint—the gap between distributed computation and sequential verbalisation—has interesting parallels to human cognition, which we explore in \cref{sec:cogni-science-perspectives}.

\paragraph{Challenges in Producing Faithful Explanations.}
\citet{TanneruEtAl2024} explored methods to steer LLMs towards faithful CoT reasoning by penalising inconsistencies during training, yet found only partial success: models often reverted to plausible-but-not-causal explanations on complex problems. This finding not only underscores the difficulty of fully eliminating LLMs’ tendency to generate plausible-but-not-causal CoT, but also highlights a concerning possibility: during fine-tuning or continual learning, models may reacquire this unfaithful reasoning behaviour, as prior work has shown that models can easily relearn previously removed concepts \cite{lo-etal-2024-large, barez2025open}. \citet{Chua2025} examined specialised ``reasoning-trained'' models and found some improvement—DeepSeek-R1 mentioned prompt hints $59\%$ of the time versus only $7\%$ for DeepSeek-V3—but still failed to acknowledge problematic influences in $41\%$ of cases. 

\citet{OpenAIReasoningMonitor2024} demonstrated that when CoT monitoring is incorporated into RL reward functions, models learn to generate benign-seeming traces while secretly executing harmful strategies—essentially ``gaming'' the CoT monitor, aligning with \citet{denison2024sycophancy}'s findings on LLMs' ability to generalise to complex gaming behaviours to achieve reward hacking. Similarly, \citet{hubinger2024sleeper} found that LLMs persist in deceptive behaviours even after safety training. While \citet{TanneruEtAl2024} showed that targeted edits can improve faithfulness in controlled settings, reasoning-trained models still omitted hidden hints in roughly $40\%$ of test cases~\citep{Chua2025}, indicating that such training interventions may offer only partial relief. %
Whether new methods can close these gaps
remains an open question.

\section{Proposed Research Directions}
\label{sec:agenda}
This section proposes research directions to help mitigate CoT unfaithfulness. In \cref{subsec:prioritised programme}, we present a prioritised research programme for developing methods and norms to assess, validate, and document CoT faithfulness. In \cref{subsec:more open problems}, we discuss additional open problems that may also be critical to addressing this issue.

\subsection{A Prioritised Research Programme}
\label{subsec:prioritised programme}
We first propose a prioritised research programme for evaluating, verifying, and reporting the faithfulness of CoT: (1) \textbf{Measurement} focuses on metrics and evaluation pipelines that can assess the severity of unfaithfulness, (2) \textbf{Activation-based validation} utilises mechanistic interpretability methods to locate and analyse the disconnect between reasoning steps and model activations, and (3) \textbf{Practice} calls for the establishment of standards and norms for documenting, stating, and sharing interpretability claims made by CoT based on the accompanying techniques in (1) and (2).

\paragraph{Measurement.} 
The simplest \emph{black-box} measurement is adversarial hint injection~\citep{Turpin2023, atanasova-etal-2023-faithfulness, siegel-etal-2024-probabilities, Anthropic2025}. We do this by injecting hints known to steer the model's final outputs and observing whether verbalised reasoning acknowledges them. The measured hint-verbalisation rates under different adversarial pressures provide a straightforward monitor of CoT faithfulness. However, hint-verbalisation rates reveal only part of the desiderata. A low rate indicates unfaithful CoT patterns, while a high rate does not guarantee all other non-hint steps are causal, nor that all causally important factors are verbalised.

Another approach for measuring CoT causality is to systematically generate alternative, counterfactual reasoning chains that omit or paraphrase individual reasoning steps that appear critical to the answer. By checking whether the model still reaches the same answer, we can quantify how often the model output changes when steps are omitted or altered~\citep {lanham2023measuring}. A low frequency suggests unfaithful reasoning patterns~\citep {lanham2023measuring}. However, deleted or paraphrased reasoning steps may produce an \emph{out-of-distribution reasoning trace}~\citep{Nostalgebraist2024} that is unlikely to arise under the model's own generation process. In such cases, the model’s behaviour may become unpredictable not because the step was causally relevant but because the edited reasoning falls outside the model’s activation distribution. Disentangling these two explanations remains an open challenge for perturbation-based tests.

We should also chart how performance on interpretability and faithfulness metrics, such as hint-verbalisation rates, changes with model size and training recipe (e.g., base vs. CoT-fine-tuned vs. RLHF). Preliminary evidence shows mixed trends: larger models may generate more faithful explanations in typical QA~\citep{siegel2025} but also hide biases more adeptly~\citep{Anthropic2025}. More studies are needed to clarify what efforts can improve faithfulness, at what cost (e.g., loss of certain capabilities), and how to push the Pareto frontier outwards~\citep{gao2025weight}.

\paragraph{Activation-based validation.} 

Measurement operates on the text of the CoT. However, the model may process multiple latent pathways in its activations when verbalising the reasoning chain, even if that chain corresponds to only one pathway. To rigorously examine the reasoning’s causal relevance and dependence, we must map each reasoning step to the relevant activation or latent pathway and test its causal role. A causal CoT is one in which the verbalised reasoning steps have corresponding activations that measurably affect later steps and the final answer. This is a looser form than faithfulness, which requires verbalising all important pathways. While a causal CoT is not necessarily faithful, it still improves on a non-causal CoT that includes irrelevant steps. Specifically, a causal CoT, while incomplete, at least ensures that the shown steps contributed to the final answer, providing partial transparency into the model's decision process.

The validation is best achieved by \emph{white-box} mechanistic interpretability methods that look into model activations. By extending mechanistic techniques such as interchange interventions~\citep{das}, activation patching~\citep{zhang2024towards}, causal tracing~\citep{Meng2022}, and circuit discovery~\citep{ameisen2025circuit} to multi-step reasoning, we can identify hidden activations tied to CoT steps and ablate them to measure their impact on final answers. This relates to the task of eliciting latent knowledge (ELK)~\citep{christiano2021wlk, mallen2024eliciting}, which aims to obtain truthful reports of information represented in a model's hidden knowledge. 

However, mechanistic validation faces several open challenges. First, interventions may cause unintended semantic shifts due to causal sensitivity~\citep{ravfogel2025gumbel}. Surgical interventions are challenging because transformers compute in a distributed manner (\cref{sec:cot-vs-internal-computation}). Second, interpretability methods that provide fine-grained assessments~\citep{zhang2024towards, Meng2022, ameisen2025circuit} often scale poorly; efficient methods such as probing~\citep{alain2017understanding, belinkov-2022-probing} and activation verbalisation~\citep{latentqa, ao} provide only coarse-grained or correlational analysis. Finally, most current methods analyse activations token-wise, while in-the-wild multi-step reasoning usually operates on spans of words. Analysing the activation dynamics that generate these tokens requires additional adaptation for most existing interpretability methods.

Because activation-based validation is costly and scales poorly, we may amortise the cost by training a \textit{verifier} model to identify non-causal CoTs in the wild~\citep{shen2026faithcotbench}. Specifically, researchers can train the verifier model on constructed CoT pairs for the same prompt: one is a causal CoT the model actually used to produce its answer, and the other is a counterfactual non-causal CoT that appears plausible but includes steps not causally responsible for the model’s decision. The task is binary classification, and the learned verifier can identify non-causal CoTs by linguistic attributes that differ from causal ones. This setup can be viewed as a \textit{prover–verifier} framework, where the model acts as a prover producing rationales, and the verifier judges their faithfulness. Success is measured by generalisation to held-out prompts to identify causal vs. non-causal CoTs correctly~\citep{Cobbe2021}. Notably, both how training-pair quality affects verifier performance and whether these verifiers learn meaningful patterns remain open questions~\citep{young2026measuringfaithfulnessdependsmeasure}.

\paragraph{Practice.} 
Beyond technical developments, we should also establish practices and standards to mitigate the spread of misplaced trust in CoT for explanatory purposes. We argue that CoT, when used for model explanations in critical domains, should always come with measurement and validation tests. However, in practice, factors like the levels of interpretability claims and downstream scenarios all affect the criticality of reporting the (estimated) faithfulness of CoT. Which tests to conduct and which guarantees to meet remain open questions.

\subsection{More Open Problems}
\label{subsec:more open problems}
Beyond the prioritised research programme, we also consider interpretable-by-design models and task-faithfulness correspondence as fruitful directions for tackling CoT unfaithfulness.

\paragraph{Interpretability-by-Design Models.}
Some prior studies have explored additional training to make LLMs faithful reasoners~\citep{TanneruEtAl2024, hase2026counterfactual}, with partial success achieved. However, transformer-based LLMs still generally cannot consistently and faithfully verbalise the pathways they use because of their distributed representations and computations. One way to bypass this limitation is to develop intrinsically interpretable models (i.e., \emph{interpretability-by-design}) whose activations and computations are much easier for humans to understand and monitor. This direction reflects \citet{rudin2019stop}'s argument that, in high-stakes domains, one should use inherently interpretable models rather than explain black-box models post hoc.

Preliminary studies have shown promise. \citet{gao2025weight} applied weight-sparsity constraints during pretraining and found that the resulting weight-sparse transformers had easier-to-identify compact circuits; \citet{elhage2022solu} proposed Softmax Linear Units (SoLU), a special activation function which imposes lateral inhibition on neurons in the same layer and token position, such that model activations become sparser and each neuron is more monosemantic; there are also studies in post-training for sparsifying attention patterns~\citep{lei2025spartan, draye2025spartan} such that calculated relations between tokens can be easily read. These efforts aim to modify the original transformer architectures or components, making them easier to analyse. Ideally, this would let us directly attribute verbalised reasoning tokens to internal computations and corresponding semantic concepts. Although an inherently interpretable model does not guarantee faithful reasoning, its transparent computations let us identify spurious reasoning steps and how they are produced. Moreover, it may offer new opportunities to align verbalised reasoning with internal pathways better, since such models have alleviated issues in vanilla transformers’ distributed computations.

However, challenges also exist. For example, an interpretability-capability trade-off often emerges: the more interpretable a model is (e.g., sparser weights or activations), the more it suffers a performance drop. To maintain a preset capability goal, we would have to scale models much larger than their dense counterparts, which is currently impractical for modern language models~\citep{gao2025weight}. More studies are needed to design better interpretability-by-design architectures and push the interpretability-capability frontier further.

\paragraph{Estimating Task-Faithfulness Correspondence.}
A model's CoT can be more causal or faithful on challenging downstream tasks than on easier ones because reasoning in complex questions is load-bearing computation (see \cref{sec:discussion}). It is thus worth exploring which tasks meet the load-bearing requirement, at which granularities (i.e., per-question or only task-wise) unfaithfulness is predictable, and the uncertainty, robustness, accuracy, and effect size of the task difference.

Knowing the task-faithfulness correspondence has two advantages. First, it can serve as an immediate, cheap prior signal to warn of potentially unfaithful CoT in practice (see \cref{subsec:prioritised programme}). Second, it indicates which tasks and scenarios may need more time and resource investment because of higher empirical unfaithfulness rates.

However, measuring the task-faithfulness correspondence can be computationally expensive. For each downstream task, we must conduct extensive causal examinations, such as the validation tests specified in \cref{subsec:prioritised programme}, for each reasoning trace in the task sample to assess its faithfulness. A large number of traces may be required to estimate the task-faithfulness correspondence for each task robustly.

\section{Alternative Views and More Discussions}
\label{sec:discussion}
In this section, we present several alternative views and broader discussions around our central claim that CoT alone, without causal validation, is not sufficient for explainability.

\begin{itemize}
  \item \textbf{CoT as useful proxy:} Some researchers emphasised that, despite fidelity issues, CoT still serves practical purposes. \textbf{(1) CoT as partial understanding,} where a reasonable explanation for a correct answer can be valuable even if it is not the exact route the model took \cite{cot-may-be-highly-informative-despite-unfaithfulness}. For example, in medical diagnosis, a model might arrive at the correct diagnosis through copying the answer from similar training samples but explain its reasoning using textbook medical knowledge. While unfaithful to internal processes, this explanation helps doctors understand and verify the diagnosis.
  However, this may lead to over-trust in high-stakes scenarios, and it could mask systematic errors that only become apparent when examining the actual computation path. \textbf{(2) CoT to improve monitorability,} where CoT has been empirically shown to be a useful monitoring signal of the system which produces the CoT~\citep{korbak2025chain, emmons2025chain, early-work-on-monitorability-evaluations, guan2025monitoring, anwar2025analyzing}. In this case, it is not essential that the CoT faithfully reflect all of the model’s internal computations—even if it does not, the monitoring system can still benefit as long as \emph{some} information about malicious intent is verbalised.
  
  \item \textbf{CoT as necessary computation in complex tasks:} Recent studies argued that CoT is %
  part of \emph{necessary computation} of models in challenging tasks~\citep{cot-may-be-highly-informative-despite-unfaithfulness, emmons2025chain, korbak2025chain}, %
  where models rely on intermediate steps to scaffold their thinking. In this case, CoT is naturally causally upstream of the final answer. %
  This resonates with CoT as a means to improve monitorability, given that some malicious behaviours require non-trivial planning~\citep{emmons2025chain}. For example, \citet{OpenAIReasoningMonitor2024} showed that CoT can reveal reward hacking behaviours in real-world reinforcement learning agents, suggesting that in complex environments, CoT can truly reflect model cognition and support effective monitoring. In this view, the usefulness of CoT stems not from its faithfulness to some latent, uninterpretable forward pass, but from the fact that it \textit{is} the forward pass: a human-legible trace of model computation. This perspective still leaves room for concern—models might alternate between using CoT for genuine reasoning and for decoration depending on the prompt or context—but suggests that CoT faithfulness should be evaluated on a task-by-task basis, rather than assumed to be always absent.

  The other indirect evidence of CoT as load-bearing computation is \emph{filler tokens}, where model performance in certain reasoning tasks can improve through the use of tokens like ``...'' or learned ``pause'' tokens that do not contribute semantically to the task but influence the model’s internal computation. For example, \citet{pfau2024let} showed that adding filler tokens enabled models to solve problems they previously failed to solve, particularly when trained with dense supervision. Similarly, appending learnable pause tokens, which can act as a type of filler token, to inputs provided a significant performance boost across several tasks~\citep{goyal2023think}. Furthermore, models trained on random or corrupted intermediate traces can perform comparably to those trained on correct reasoning paths~\citep{stechly2025semantics}. Together, these results question what proportion of improvement from CoT is due to an additional (possibly meaningless) token-based computation rather than human-like step-by-step reasoning~\citep{pfau2024let}.
  
  \item \textbf{Scaling and better training will bridge the gap.} Some believe that CoT unfaithfulness will naturally diminish as models become more capable. According to this view, as models improve in overall reasoning capabilities, the gap between their internal computation and verbalised explanations should narrow. For instance, larger models have shown improved performance on complex reasoning tasks~\citep{Bubeck2023}, and specialised training techniques like reinforcement learning have demonstrated some success in making models more honest about their reasoning~\citep{Anthropic2025}. However, this view faces several challenges: (1) we lack clear evidence that larger models produce more faithful explanations rather than just more plausible ones, and on the contrary, there is evidence that models produce less faithful explanations as they get larger \cite{lanham2023measuring}, (2) 
  there is no evidence that training methods that enhance task performance also incentivise models to produce more faithful explanations, and (3) the architectural constraints of transformers, such as distributed processing, may fundamentally limit how well they can verbalise their internal computations (\cref{sec:cot-vs-internal-computation}). Recent work suggested that more advanced models may simply become better at hiding their unfaithfulness, making it harder to detect when explanations diverge from actual computation \cite{OpenAIReasoningMonitor2024}.

  \item \textbf{Assisting CoT with future interpretability tools:} 
  Proponents argued that improvements to interpretability techniques, such as activation patching and attention visualisation, could provide complementary insights into model computation \cite{Burns2022}. For example, while a model's CoT might not fully capture its reasoning process, these tools could help identify which parts of the input influenced the output and how different model components contributed to the final decision. This approach has the advantage of working with existing models and could provide more insights than CoT alone. However, current interpretability tools face significant limitations: they often require extensive compute and do not scale to frontier models, and can be difficult to interpret even for experts \cite{bereska2024mechanistic}. Moreover, these tools typically provide post-hoc analysis rather than real-time explanation, making them less practical for real-world deployment. 
  \item \textbf{Is faithfulness an overly strict criterion for CoT?} 
  Human reasoning can also exhibit unfaithful rationalisation (see \cref{sec:cogni-science-perspectives}). This raises a question: are we setting an overly stringent standard for CoT as an explanatory tool, given that we normally do not require alignment between humans' reasoning and their actual thoughts? We believe the answer is \emph{no}, for three reasons: (1) Humans' thoughts in critical domains can be cross-examined with direct evidence such as official documents and licensing. At the same time, CoT is often used without such cross-examination (see \cref{sec:background}), which is why we argue for either pairing CoT with additional verification or developing faithful CoT. (2) Humans normally are not trained to separate their real thoughts from verbalised reasoning. At the same time, language models can be deliberately or inadvertently optimised to hide hidden objectives in activations without verbalising them~\citep{OpenAIReasoningMonitor2024}. (3) Legally and morally, we cannot arbitrarily inspect humans' real thoughts (e.g., by electroencephalography or lie detector tests), while normatively we can inspect models' activations by mechanistic interpretability approaches; this makes aligning CoT with internal computations a feasible technical goal. Altogether, these reasons justify faithfulness as a reasonable standard for CoT as an explanatory tool, though this seems higher than what we expect of humans.
  
\end{itemize}

In summary, these perspectives complement our central claim that CoT needs to be faithful for explanatory purposes and explain why we do not reject CoT for explanations altogether. CoT validated by other interpretability methods may provide useful information and, in some complex tasks, may be more causally involved in producing the answer. Thus, CoT alone may be insufficient for explanation, but it can still serve as a useful explanation, or part of one, when appropriately validated.

\section{Conclusion}
CoT is widely viewed as a step towards interpretable language models. Yet we argue that this promise is not yet fulfilled. CoTs can appear convincing, while not faithfully reflecting the true decision process of the model. However, currently CoT is over-trusted in downstream application domains.
Although we agree that CoT can be a useful mechanism for eliciting reasoning traces from models, it should not be mistaken for ground truth without causal validation. We then propose a framework to evaluate CoT faithfulness and discuss empirical and architectural drivers of unfaithfulness. 
Going forward, we recommend that researchers and practitioners (1) avoid treating CoT as sufficient evidence of interpretability, (2) adopt more rigorous validation methods, and (3) accompany interpretability claims based on CoT with the evidence from (2). We refer readers to \cref{sec:impact statement} for a discussion of the social impacts and ethical concerns associated with establishing the notion of CoT unfaithfulness.

\section*{Acknowledgements}
We thank Christopher Summerfield, Erik Jenner, Peter Hase, Jacob Pfau, and Sören Mindermann for thoughtful comments and discussions that shaped the project, and some of the framing.

We also thank:

Elliot Fosong, Isaac Friend, Shoaib Ahmed Siddiqui, Itay Yona, Usman Anwar, Stephen Casper, Nicky Pochinkov, Antonio Valerio Miceli Barone, Shoana Gaosh, Aidan O’Gara, Jialin Yu, James Oldfield, Lovis Hendrich, Mor Geva, Itay Izhak, Zheng Zhao, Aviv Ovadya, Hasan Sajid, Philip Quirke, Adi Simhi, Narmeen Oozeer, Michelle Lo, Eric Sun, Rudolf Laine, and David Manheim for their comments on the initial set of ideas.

\small{
\bibliography{reference}
}

\newpage
\appendix
\crefalias{section}{appendix}
\DoToC
\newpage
\section*{Appendix}
\section{Does Chain-of-Thought Mirror Human Reasoning Patterns?}
\label{sec:cogni-science-perspectives}

Interestingly, the disconnect between a verbalised explanation and the underlying computational process
is not unique to artificial networks. While we do not claim that LLMs think like humans or have human-like cognition, and caution readers against anthropomorphising AI models~\cite{placani2024anthropomorphism}, cognitive psychology and neuroscience have documented \emph{conceptually similar} phenomena in humans that offer both cautionary analogies and potential inspirations for understanding and improving AI explanations:

\paragraph{Confabulation and Post-Hoc Rationalisation.}
\citet{Nisbett1977} demonstrated that people often provide plausible but inaccurate explanations for their decisions. In many cases, humans are unaware of the true drivers of their behaviour and instead offer confabulated narratives. This suggests that apparent step-by-step explanations 
(e.g., a person explaining their choice of a product by listing logical factors, when in reality they were influenced by subtle environmental cues) can be mere rationalisations, much like an LLM's CoT may justify an answer without revealing its true genesis.

\paragraph{The Left-Brain Interpreter.}
Classic split-brain studies revealed that the brain's language-dominant hemisphere will generate explanations for actions initiated by the opposite hemisphere---even when it lacks access to the true cause \cite{Gazzaniga1989}. This ``left-brain interpreter'' continuously fabricates a coherent story, masking the distributed and parallel nature of neural processing. 
While we acknowledge this is a speculative analogy, recent work on distributed computation in transformers \cite{Dutta2024, nikankin2025arithmetic, quirke2024} suggested that LLMs may similarly generate sequential narratives that mask their parallel computational processes.

\paragraph{Parallel Processing and Sequential Narratives in Humans and LLMs.}

The human brain operates via distributed, parallel processes yet yields a sequential subjective narrative of perception and reasoning (e.g., we experience a continuous stream of consciousness despite parallel neural processing). Predictive processing theories posit that the brain constantly generates hypotheses about incoming information and updates its internal model to minimise prediction errors \cite{Clark2013}. The conscious narrative we experience is a simplified summary of this complex process. 
While speculative, we note that an LLM's CoT could be viewed as one possible narrative path sampled from its latent distributed computations. 
Notably, the brain's narrative can be adaptive: if predictions strongly contradict reality, error signals prompt a revised understanding. This hints that an LLM might benefit from a similar mechanism to check and adjust its CoT when steps conflict with its latent knowledge, though as noted in \cref{sec:cot-unfaithfulness}, models already exhibit some internal error correction.

\paragraph{Metacognition and Error Monitoring.}
Humans exhibit metacognition: the ability to reflect on and evaluate their thoughts. The brain even has dedicated circuitry for error monitoring: for example, the anterior cingulate cortex emits error-related signals when a mistake or conflict in reasoning is detected \cite{Botvinick2001,Yeung2004}. These signals can trigger heightened attention or strategy adjustment, preventing us from confidently persisting in a flawed line of thought. While models already show some internal error correction (as discussed in \cref{sec:cot-unfaithfulness}), explicit metacognitive mechanisms might help improve the faithfulness of their verbalised reasoning by making the correction process more transparent.

\paragraph{Towards Self-Correcting Narratives (Predictive Coding in AI).}
In neuroscience, predictive coding provides a powerful model of how brains correct their narratives by minimising surprise. While models already show some ability to plan and correct errors internally, we could potentially enhance this by designing an LLM reasoning process that explicitly forecasts the likely outcome of its current chain-of-thought and compares it to the model's actual next decisions or the final answer.

\paragraph{Dual-Process Reasoning and System-2 Analogues.}
Cognitive science often distinguishes between fast, intuitive thinking (System 1) and slow, deliberative reasoning (System 2) \cite{Kahneman2011}. Humans can sometimes engage the latter to double-check or override the impulses of the former. While today's LLMs do not have a clear architectural separation between intuitive and logical processing, they do show different behaviours in different contexts, sometimes answering directly and sometimes engaging in step-by-step reasoning. 
This has led some researchers to speculate about architectures that explicitly incorporate a System-2 module for reasoning. For example, \citet{Bengio2017} proposed a ``consciousness prior'' for neural networks, encouraging a sparse, sequential activation of neurons corresponding to something like conscious thought.

These cognitive analogues may offer useful perspectives for examining CoT faithfulness and suggest potential directions for improving it. Just as humans benefit from metacognitive awareness and error monitoring, future LLMs might incorporate more explicit self-checking mechanisms. However, implementing such systems faces similar challenges to human cognition: the monitoring system could be as fallible as the process it monitors. 
The key insight from cognitive science is that narrative construction---whether human or artificial---inherently simplifies complex parallel processes into sequential stories.

\section{Additional Experimental Results}
\label{sec:additional results}

\subsection{Human Verification}
To assess the reliability of our automated pipeline, we manually classify 100 randomly sampled CoT-centric papers without seeing the model's labels. The resulting agreement rate within one level is 80.0\%, with a false positive rate—cases where we label a paper as L0 or L1 but GPT-4o labels it as L3 or L4—of 14.3\%. Further investigation shows that most disagreements arise from papers that use CoT within their methods (e.g., as a reward signal or as a target for pruning), where faithful CoT may be preferred in these settings but no direct interpretability claim is made (however IET-related words may be mentioned, such as ``explanation'', ``monitor'', ``audit'', etc.). In this case, GPT-4o still tends to assign these papers L3, whereas the annotator labels them L0, because our rubric focuses on explicit IET claims. Of the 10 papers GPT-4o labels L3/L4 but the annotator does not, 9 are labelled L0 by the annotator for this reason.

\subsection{Reverse-Direction Analysis}
Following the pipeline in \cref{fig:detect_pipeline}, we conduct a reverse-direction analysis by first collecting 276 arXiv papers that make interpretability a central goal and then classifying them based on the roles they assign to CoT. The results are in \cref{tab:reverse-analysis}. Notably, among the 276 interpretability-centric papers, 20 (7.2\%) adopt CoT as the interpretability mechanism itself, and 81 (29.3\%) incorporate it as a component of the proposed pipeline.

\begin{table}[t]
\centering
\small
\caption{Role of CoT in 276 recent papers with interpretability as a central goal.}
\label{tab:reverse-analysis}
\begin{tabular}{lrr}
\toprule
\textbf{Role of CoT} & \textbf{Papers} & \textbf{\%} \\
\midrule
A. Absent                              & 165 & 59.8 \\
B. Related work only                   &   5 &  1.8 \\
C. Baseline only                       &   0 &  0.0 \\
D. Component of proposed method        &  81 & 29.3 \\
E. Interpretability mechanism itself   &  20 &  7.2 \\
F. Object of study                     &   5 &  1.8 \\
\midrule
Total                                  & 276 & 100.0 \\
\bottomrule
\end{tabular}
\end{table}

\section{Prompt Templates for the Classification Pipeline}
\label{sec:prompt}
Here we present the system and user prompts used in the analysis pipeline. We embed the relevant excerpts and the paper's title into the user prompt. The user and system prompts form a complete message to the classification model. We then parse the returned output to acquire this paper's predicted level.

\lstdefinestyle{promptstyle}{
  basicstyle=\ttfamily\fontsize{6.7}{8.2}\selectfont,
  columns=fullflexible,
  breaklines=true,
  breakatwhitespace=false,
  postbreak=\mbox{\textcolor{gray}{$\hookrightarrow$}\space},
  breakindent=1.5em,
  upquote=true,
  showstringspaces=false,
  keepspaces=true,
  aboveskip=0pt,
  belowskip=0pt,
}
\lstdefinestyle{schemastyle}{
  style=promptstyle,
  language=Python,
  keywordstyle=\color{NatureRiver},
  stringstyle=\color{NaturePine},
  commentstyle=\color{gray}\itshape,
}
\newtcblisting{promptbox}[2][]{
  enhanced, breakable, listing only,
  colback=#2!4!white, colframe=#2!85!black,
  coltitle=white, fonttitle=\bfseries\small,
  boxrule=0.6pt, arc=2pt, left=4pt, right=4pt, top=9pt, bottom=5pt,
  attach boxed title to top left={xshift=8pt, yshift=-\tcboxedtitleheight/2},
  boxed title style={colback=#2!85!black, colframe=#2!85!black, arc=2pt, boxrule=0pt},
  before skip=10pt, after skip=10pt,
  #1
}

\begin{promptbox}[title={System prompt}, listing options={style=promptstyle}]{NatureRiver}
You are a careful research-methods analyst auditing how AI papers talk about chain-of-thought (CoT).
Judge ONLY from the excerpts provided. Never infer from the title, the venue, or what such a paper
usually does. If the excerpts do not settle a field, choose the most conservative value and set
confidence to "low". Every quote you output must appear verbatim in the excerpts.
\end{promptbox}

\begin{promptbox}[title={User prompt}, listing options={style=promptstyle,
    emph={BACKGROUND,TASK,DECISIVE,COUNTERFACTUAL,BOUNDARY,RULES,OUTPUT,PAPER,TITLE,EXCERPTS},
    emphstyle={\color{NatureRiver}},
    moredelim={[is][\color{NatureClay}]{@}{@}}}]{NaturePine}
BACKGROUND. A CoT trace is text a model emits before its answer. A growing literature shows these
traces can be unfaithful: they may not reflect the computation that actually produced the answer.
Papers nevertheless invoke CoT as an interpretability, explainability, or transparency (IET) device.
Your job is to separate papers where this is loose rhetoric from papers whose substantive claims
DEPEND on traces being faithful.

TASK. Assign exactly one load-bearing tier, using the highest tier that applies.

L0 NO CLAIM - no IET language about CoT anywhere in the excerpts.
L1 RHETORICAL - IET language appears only as framing: an adjective in the abstract, motivation in the
   introduction, a related-work summary. Decisive test: delete every IET word and no contribution,
   method, experiment, or conclusion of the paper changes.
L2 PRESENTATIONAL - CoT traces are surfaced to a human as explanation or justification (qualitative
   examples, user-facing rationales, "our system explains its decisions"), but no downstream component
   or claim treats the trace as evidence about the model's internal computation.
L3 METHODOLOGICALLY LOAD-BEARING - a technical component treats the trace as a valid account of the
   model's process. Examples: diagnosing model errors by reading traces; rewarding or verifying
   individual reasoning steps; filtering, selecting, or distilling training data on the basis of
   "reasoning quality"; a proposed metric that claims to measure WHY the model answered; an
   "interpretable-by-design" system whose interpretability consists of emitting a CoT.
L4 SAFETY/DEPLOYMENT LOAD-BEARING - the trace is relied on for monitoring, auditing, human oversight,
   detection of deception, hallucination, or misalignment, a regulatory or compliance argument, or a
   claim that a system is safe or trustworthy enough to deploy in a high-stakes domain (medical, legal,
   financial, autonomous) because its reasoning is visible.

DECISIVE COUNTERFACTUAL. Before choosing between L1/L2 and L3/L4, ask: if it were proven tomorrow that
CoT traces do not reflect the model's computation, which specific claims in this paper become false?
If the honest answer is "none, the prose just reads worse", the tier is L1 or L2. If a method,
metric, safety argument, or deployment recommendation collapses, the tier is L3 or L4.

BOUNDARY RULES.
- CoT used purely to improve task accuracy, with no IET language, is L0 - however elaborate the method.
- "Step-by-step", "structured", "decomposed", "transparent pipeline" describing a SYSTEM's architecture
  rather than the model's reasoning is not an IET claim about CoT. Do not count it.
- A paper that measures CoT faithfulness, or argues traces are unfaithful, is NOT load-bearing on
  faithfulness being true. Tier it by any OTHER claims it makes, and set faithfulness_engagement to
  "empirically_tested".
- Human-written or dataset-provided rationales that are never claimed to reflect the model's own
  computation are at most L2.
- One qualifying sentence is enough for L1; L3 and L4 require an actual component, metric, or
  recommendation, which you must name in counterfactual_break.

OUTPUT. Return one JSON object and nothing else:

{
  "tier": "L0" | "L1" | "L2" | "L3" | "L4",
  "makes_iet_claim_about_cot": true | false,
  "claim_locus": ["abstract" | "intro" | "related_work" | "method" | "experiments" | "discussion" | "safety_or_deployment"],
  "trace_consumer": "none" | "human_reader" | "end_user" | "automated_component" | "paper_metric",
  "faithfulness_engagement": "absent" | "mentioned_in_passing" | "cites_unfaithfulness_work" | "caveated_as_limitation" | "empirically_tested",
  "domain": "general_nlp" | "medical" | "legal_finance" | "autonomous_robotics" | "education" | "code_software" | "agents_tools" | "multimodal" | "science_math" | "other",
  "counterfactual_break": "<one sentence naming the component/claim that fails if traces are unfaithful, or \"\" for L0/L1>",
  "evidence": [{"quote": "<verbatim>", "section": "<abstract|intro|method|...>"}],
  "confidence": "high" | "medium" | "low",
  "reasoning": "<<=60 words justifying the tier, referring to the counterfactual>"
}

PAPER TITLE: @{title}@

EXCERPTS:
@{pack}@
\end{promptbox}

\section{Impact Statement and Ethical Concerns}
\label{sec:impact statement}

This paper argues that chain-of-thought (CoT) should not be treated as reliable evidence of AI interpretability. One potential positive impact of this work is that researchers and practitioners may adopt stronger evaluation practices to distinguish verbalised explanations from causal evidence about a model's underlying computations. This could reduce the risk of trusting deployed non-transparent systems in scenarios where transparency is critical.

One potential negative impact of studies that establish or reinforce the notion of CoT unfaithfulness, including this work, is that widespread recognition that CoT explanations may be unfaithful could give model developers reasons to disclaim legal or social responsibility for misleading outputs. This concern is not entirely new. Broader awareness of model hallucinations has already led AI services such as ChatGPT and Claude to display warnings that their outputs may contain mistakes. However, we believe these warnings are, overall, beneficial for end users. More concretely, informing users about the possibility of CoT unfaithfulness lets them choose a more informed level of trust in a model's explanation. Without this awareness, users may discover these explanations' limitations only after a consequential failure. Moreover, greater public awareness of CoT unfaithfulness establishes it as a foreseeable risk that, in fact, imposes transparency, accountability, and explainability obligations on model providers~\cite{euaiact2024, gpaicop2025}, requiring them to respond to and address it, for example by (1) disclosing the possibility of CoT unfaithfulness so that users can calibrate their reliance appropriately and (2) developing models that can produce more faithful and aligned reasoning traces; both measures benefit users relative to a setting in which CoT explanations are presented without qualification. In summary, awareness of CoT unfaithfulness---or any studies that identify and substantiate this concern, including this work---should not be taken to absolve model developers of their legal or social responsibilities; regulatory efforts~\cite{euaiact2024, gpaicop2025} increasingly place responsibility on providers to identify, disclose, and mitigate such risks. However, a full account of the legal liability of AI companies and users under different conditions, such as whether the CoT is causally relevant or whether users are misled by the explanation, falls outside the scope of this work, and we consider it an important direction for future research.

\end{document}